# Towards a Normative Theory of Scientific Evidence - a Maximum Likelihood Solution*

David B. Sher
Computer Science Dept.
SUNY at Buffalo
Buffalo NY 14260
sher@cs.buffalo.edu — sher%cs.buffalo.edu@ubvm.bitnet

June 6, 1990

## Abstract

A scientific reasoning system makes decisions using objective evidence in the form of independent experimental trials, propositional axioms, and constraints on the probabilities of events. As a first step towards this goal, we propose a system that derives probability intervals from objective evidence in those forms. Our reasoning system can manage uncertainty about data and rules in a rule based expert system. We expect that our system will be particularly applicable to diagnosis and analysis in domains with a wealth of experimental evidence such as medicine. We discuss limitations of this solution and propose future directions for this research. This work can be considered a generalization of Nilsson's "probabilistic logic" [Nil86] to intervals and experimental observations.

*We gratefully acknowledge the work of Carolyn Sher in editing this work and the assistance of Moises Sudit regarding nonlinear programming.

## 1 Introduction

Expert systems were originally defined as a method of encoding the knowledge and reasoning of a human expert into a computer program. However, for many diagnostic and analytical reasoning problems we would prefer that the computer program base its reasoning on objective facts rather than human expertise, especially when no experts are available or the reasoning of the available experts is suspect. Thus, we propose a system for reasoning from objective criteria.

When objective data is collected by scientists and engineers it is often in the form of independently distributed trials or experiments. An example of experimental data is: in a sample of 100 toys 10 arrived damaged. Some of these experiments verify facts, as in our example, others may verify rules, i.e. of 10 red toys 1 arrived damaged, establishing a relationship between redness and damage. Several rules can conflict, i.e. of 10 toy trucks 9 arrived damaged which raises the question of the red toy truck. This paper describes a computationally feasible algorithm for making decisions from this type of evidence. We present a preliminary methodology for evidence combination, where the source of the evidence is experimental in nature.

Our scheme can also account for other methods of expressing objective information such as axiomatic statements, ie. trucks have wheels, and restrictions on the probabilities of events, ie. the probability of heads on a fair coin is 0.5, the probability of heads when Joe is flipping the coin is somewhere between 0.6 and 0.8. Initially we will focus on our management of experimental evidence since this is the unique feature of our approach.

## 2    Previous Work

Our work was inspired largely by Henry Kyburg and Ron Loui's work [Kyb87, Lou84] at the University of Rochester on the logical foundations of statistical inference. Their work uses experimental results to derive sets of confidence intervals and then develops an algebra of these sets of confidence intervals.

Other important influences on this work is the Dempster-Shafer system [WH82, WS88] for managing uncertainty and the analyses thereof [Gro85, HL85, Kyb87]. Dempster-Shafer reasoning expresses belief as mass functions over sets of possible beliefs and develops a calculus of such mass functions for evidence combination. While such mass functions may be a good way of summarizing human expertise, we felt that they were not suitable for expressing the uncertainty inherent in objective evidence. Hau and Kashyap [HK87] suggested a method of applying Dempster-Shafer and fuzzy reasoning to rule based expert systems which we consider one of the best approaches to this problem if the rules and facts in the system are derived from a subjective or qualitative source.

This work owes a great deal to the Bayesian interval approach to uncertainty; it reduces to Good's Bayesian interval approach [Goo50, Goo83] when only interval probabilities are supplied. Our work also is meant to handle the cases the same class of applications as Judea Pearl's probability networks [Pea86a, Pea85] and other Bayesian network paradigms [Lev86, Sha85, Pea86b, ED84, Mar85, CR87, CC87, She89]. However, we do not have to supply or calculate precise probabilities and can take into account the fact that certain calculations are much better informed than others.

Our work generalizes Nilsson's "probabilistic logic" [Nil86] where he introduces the JDVs (with a different name). His work analyzes the case where all the information are point probabilities for events; our work reduces to his when our information is in this form.

## 3    Definitions

The input to our system consists of a set of propositions, and a set of experiments that test logical combinations of these propositions. For example, consider playing poker against Harry. What can we deduce from Harry's body language about his hand? Consider these two propositions:

1. A: Harry lit his pipe.

2. B: Harry has 2 pair.

Some experiments about the relationship between these two events are: (1) In the first 30 hands Harry lit his pipe in 9 of them. (2) In the next 40 hands Harry had two pair in 5 of them. (3) In the following hands you noticed that of the 6 times he lit his pipe 5 of those times he had two pair.

The joint distribution of events in our domain contain all the information about the domain that is useful for deduction. The joint distribution is the probability distribution over the elements of the truth table, i.e. in our poker example:

- $a$: A and B

- $b$: A and not B

- $c$: not A and B

- $d$: not A and not B

If we knew the probability of $a, b, c$, and $d$, we could determine the probability of any logical combination of events and all the conditional probabilities too. Hence, to engage in deduction from experimental evidence, we will study the issue of estimating a joint distribution from experimental evidence.

# 4 Maximum Likelihood Estimation

In this work we propose estimating the probability of a proposition as being one of the probabilities derived from a joint distribution that maximizes the probability of the experimental results.

In estimation theory if we are estimating the value of $e$, the function that maps values of $e$ into probabilities of the observed data is called *the likelihood function*. The value of $e$ that maximizes the likelihood function is called the *maximum likelihood estimate*. The maximum likelihood joint distribution is the joint distribution that maximizes the probability of the experimental results.

We assume that each experiment is identically independently distributed (iid) and that the sampling was unbiased. The probability of observing a logical combination of the primitives (such as "A or B") is a sum of probabilities in the joint distribution; in our poker example the probability of A is $a + b$ and B is $a + c$. Thus the probability of an experiment on A where A was observed 3 out of 5 times is then $(a + b)^3(c + d)^2$. The iid assumption implies that the probability of a set of experiments is the product of the probabilities from each experiment. Hence, the likelihood function for statements (1) and (2) of our poker example is:

$$(a + b)^9(c + d)^{21}(a + c)^5(b + d)^{35}$$

Conditional experiments, where the result is only reported when a condition applies ie: statement (3), have a conditional probability. The probability of a conditional experiment is the ratio of the probability of the conjunction of the condition and the event and the probability of the condition. Thus the probability of statement (3) is

$$\frac{a^5 b}{(a + b)^6}$$

The probability of (1), (2) and (3) is

$$a^5 b(a + b)^3(c + d)^{21}(a + c)^5(b + d)^{35}$$

Since we had an experiment on A where A occurred 9 times, we could perform up to 9 conditional experiments on A and still express the likelihood function as a polynomial (rather than a rational function). This paper's analysis is restricted to sets of experiments whose likelihood function is a polynomial; this means that given N experiments using condition X, an experiment was done where condition X occurred at least N times.

# 5 Observation Space

Each joint distribution is an assignment of probabilities to a finite set of mutually exclusive events; hence a joint distribution can be considered a vector (JDV) and the locus of joint distributions is a set of points in a vector space. In our poker example there are four mutually exclusive events, $a, b, c$ and $d$; assignments of probabilities to these events correspond to four dimensional vectors. The probability of an observation is a linear function of the JDV, in our poker example A has a probability of $a + b$. These linear functions define a dual space of observation vectors (OV) with the same dimensionality, whose coefficients are 1 if the corresponding element of the joint distribution is compatible with the observation and 0 otherwise. In our poker example A is represented by the vector $(1, 1, 0, 0)$ and B is $(1, 0, 1, 0)$ and A but not B is the vector $(0, 1, 0, 0)$. The probability of an observation under a joint distribution is the dot product of the OV with the JDV.

A set of observations, resulting from experiments, corresponds to a set of OV's that span a vector space called the *observation space.* This space will often be lower dimensional than the space spanned by JDV's. The space spanned by the JDV's is the cross product of the observation space and the space spanned by vectors perpendicular to the observation space. We call the space spanned by vectors perpendicular to the observation space the *null space.*

Note that if the difference of two JDV's is an element of the null space, for a set of experiments, then the probability of each observation made in the experiments is the same for both JDV's. Since the probability of all the observations is the product of the probabilities of each observation, the probability of the results from the set of experiments is the same for both JDV's.

Hence, the results from those experiments can not tell us which of the corresponding joint distributions is a better model. For example, consider the evidence from statements (1) and (2) of the poker example; the observations correspond to these four vec-

tors:

| Proposition | Vector |
|---|---|
| A | $(1, 1, 0, 0)$ |
| not A | $(0, 0, 1, 1)$ |
| B | $(1, 0, 1, 0)$ |
| not B | $(0, 1, 0, 1)$ |

; these vectors span a three dimensional space. The vector $(1, -1, -1, 1)$ is perpendicular to all four of these vectors. The space spanned by this vector is the null space of any set of observations where A, not A, B and not B were observed. Any two JDV's that differ by an element of the null space (such as $(0.25, 0.25, 0.25, 0.25)$ and $(0.5, 0, 0, 0.5)$) yields the same likelihood for any set of observations; the observations made in statements (1) and (2) yield a likelihood of $0.5^{70}$.

# 6 Concavity and Uniqueness of the Maxima

Two different joint distributions maximize the likelihood of a set of experiments only if their JDV's differ by an element of the null space for that set of experiments. Hence if we can find a single maximum likelihood joint distribution we know what all of them look like.

First, we show that the likelihood function, $L$, is concave. Assume that $j_1$ and $j_2$ are two JDV's; let $f(\gamma) = \log L(j_1 + \gamma(j_2 - j_1))$; $L(x) = \prod_i s_i(x)$ where $s_i$ are sums of elements of joint distribution. Each $s_i$ is the dot product of an observation vector and a JDV; hence $L = \prod_i o_i \bullet (j_1 + \gamma(j_2 - j_1))$.

$$f(\gamma) = \sum_i \log(o_i \bullet (j_1 + \gamma(j_2 - j_1)))$$

$$\frac{d^2 f(\gamma)}{d\gamma^2} = -\sum_i \frac{(o_i \bullet (j_2 - j_1))^2}{(o_i \bullet (j_1 + \gamma(j_2 - j_1)))^2}$$

Since the second derivative of $f$ is the arithmetic inverse of a sum of squares, it is never positive; hence $f$ is concave. Since $j_1$ and $j_2$ are arbitrary JDV's $L$ must be concave.

Now assume that $j_1$ and $j_2$ are maximizers of $L$. Since the log likelihood function is concave, all $j_1 + \gamma(j_2 - j_1)$ with $0 < \gamma < 1$ also maximize the joint likelihood function. Hence, the second derivative of $f$ is 0 for $0 < \gamma < 1$ because $f(\gamma) = \log L(j_1 + \gamma(j_2 - j_1))$ is a constant function.

$$0 = \frac{d^2 f(\gamma)}{d\gamma^2} = -\sum_i \frac{(o_i \bullet (j_2 - j_1))^2}{(o_i \bullet (j_1 + \gamma(j_2 - j_1)))^2}$$

Clearly $\forall_i o_i \bullet (j_2 - j_1) = 0$; hence $j_2 - j_1$ is in the null space of our observations since it is perpendicular to all of our observation vectors. If $j_1$ is a maximum likelihood JDV then $j_2$ is a maximum likelihood JDV if and only if $j_1 - j_2$ is in the null space.

An important feature of our system is — introducing a new observable O with experiments in which O was true $p$ of the time causes our system to assign a probability of $p$ to O; this is in accordance with intuition. This feature derives from the fact that the $\beta(n, m)$ distribution is maximized at $\frac{n+1}{m+1}$.

# 7 Computational Methods

To discover a vector of positive numbers whose components sum to 1 which maximizes the likelihood function is a problem in nonlinear programming. Constraining the vector to be a probability distribution limits the search space to a convex bounded set (in the form of a hypertetrahedron). Thus we suggest applying the method of feasible directions of Topkis and Veinott [BS79]; this algorithm provably converges to a maximum of the polynomials we propose here.

Given a maximum likelihood joint distribution, the method of feasible directions can also discover the bounds for any proposition. If a proposition has been directly experimented with, ie. proposition A in our poker example, its probability is the same in all likelihood maximizing distributions. If no experiments have been done on a proposition, such as not A and not B in the poker example, upper and lower bounds for its probability in likelihood maximizing joint distributions can be determined by the method of feasible directions.

We believe that the average computational cost of our system is proportional to the number of propositions whose probabilities are bounded and on which experiments have been performed; hence entering and retrieving information is linear in the amount of information entered.

# 8 Propositional Axioms

Statements of propositional logic, like A→B can be added to this system. They just fix the probability of certain elements of the joint distribution at 0. If we added B→A to our poker example then the probability of $c$ would be known to be 0 and the likelihood function would be: $a^{10}b(a + b)^3 d^{21}(b + d)^{35}$. If experimental evidence directly contradicts an axiom, either the evidence or the axiom must be discarded.

In our system, the probability of a proposition can be limited to a specified range. This is a linear constraint on the values of the JDV's. Since the space of legitimate JDV's is the intersection of linear constraints the method of feasible directions still applies. Thus we can insert into our system experimental evidence, axiomatic knowledge, and probability intervals.

# 9 Simple Example

In our poker example (0.0375, 0.2625, 0.0875, 0.6125) is a JDV for statements (1) and (2); we have discovered using numerical techniques that the maximum likelihood JDV for statements (1), (2), and (3) is approximately (0.174,0.066,0,0.76). Since the null space for (1),(2) and (3) is itself null this is the only maximum likelihood JDV. From these JDV's we can compute maximum and minimum probabilities for these events:

| Event | (1) and (2) Min :Max | (1),(2) and (3) |
|---|---|---|
| A | 0.3   :0.3 | 0.24 |
| B | 0.125:0.125 | 0.174 |
| A and B | 0   :0.125 | 0.174 |
| B given A | 0   :0.417 | 0.725 |
| B given not A | 0   :0.179 | 0 |
| A given B | 0   :1 | 1 |
| A given not B | 0.2   :0.343 | 0.080 |

This table demonstrates several facts about our poker example:

- In the case where only observations (1) and (2) have been made:

    - The probabilities for the statement A and the statement B are appropriate given the evidence.

    - Any level of correlation between A and B is possible.

    - B being false substantially constrains A.

    - Harry's lighting his pipe tells you very little about his hand.

- In the case where observations (1) and (2) and (3) have been made:

    - Since A and B are connected by statement (3) their probabilities are brought closer together.

    - B→A.

    - If Harry is lighting his pipe definitely bet on his having two pair.

 - If Harry isn't lighting his pipe then bet your fortune he doesn't have two pair (this is one of the difficulties with this approach).

Our poker example demonstrates how our system balances general knowledge about the frequency of an event with inferential knowledge about an event.

To further examine this case consider what happens if we observe another 200 hands in which Harry never has two pair (4). The JDV for this case is (0.0396257, 0.130458, 0, 0.829916). We can update our table to include this assertion as below:

| Event | (1) and (2) Min :Max | (1),(2) and (3) | (1)-(4) |
|---|---|---|---|
| A | 0.3   :0.3 | 0.240 | 0.170 |
| B | 0.125:0.125 | 0.174 | 0.040 |
| A and B | 0   :0.125 | 0.174 | 0.040 |
| B given A | 0   :0.417 | 0.725 | 0.233 |
| B given not A | 0   :0.179 | 0 | 0 |
| A given B | 0   :1 | 1 | 1 |
| A given not B | 0.2   :0.343 | 0.080 | 0.136 |

Statement (4) has reduced the probability of B considerably even in the case where A is true. Harry's lighting his pipe is still a very good clue that he may have two pairs.

# 10   Inadequacies and Improvements

There are three major difficulties with this system:

1. It sometimes behaves in an unintuitive fashion;

2. When convenient, it assigns the probability of 0 to events;

3. It is not consistent with Bayesian conditionalization.

The initial problem will occur in any normative system since there are cases where the correct decision is an unintuitive one. Such probability paradox's abound and can be very subtle. Thus human judgment of intuitive systems is not useful for evaluating normative probability systems. A better judgment of its effectiveness is to evaluate the effectiveness of expert systems built using this system.

The application of the maximum likelihood principle leads to the second problem. Assuming certain events are impossible often maximizes the likelihood of a set of observations (if the *impossible* event is not observed). This leads to the system making overly strong statements such as saying that patients without symptoms do not have colds under observations [3].

The maximum likelihood principle also yields the last problem. If one wants to add into our system new information, then using Bayesian updating will not generate the probabilities that adding the information directly into the system and updating its constraints or polynomial would.

The last two problems can be eliminated by discarding the maximum likelihood principle, instead, using the likelihood function and Bayes' law to translate a prior probability distribution over JDV's into a posterior distribution of JDV's. Then the probability of any event or combination of events is computed by integrating over the posterior distribution of JDV's. This integration can be speeded by the fact that most probable JDV's in this distribution will lie near a maximum likelihood JDV's.

We are investigating deriving the distribution over joint marginal distributions with Occam's razor. The probability of a JDV would be a function of its simplicity. Solomonoff has developed methods for evaluating the simplicity of distributions and assigning probabilities based on this evaluation [Sol89].

## 11  Conclusion

We have proposed a computationally efficient method for propositional evidence combination given logical axioms, point probabilities, probability intervals and experimental evidence. Our system returns probability intervals that are often point probabilities; it follows a strictly Bayesian interpretation of the evidence subject to the maximum likelihood principle. In domains where the evidence is largely objective such as medical diagnosis or computer vision such a system may be superior to those based on Dempster-Shafer reasoning or probability networks.

# Session 10:

## Non-Monotonic Reasoning and Conflicting Data